# An Anytime Algorithm for Decision Making under Uncertainty

**Michael C. Horsch**
horsch@cs.ubc.ca

**David Poole**
poole@cs.ubc.ca

Department of Computer Science
University of British Columbia
2366 Main Mall,
Vancouver, B.C., Canada V6T 1Z4

## Abstract

We present an anytime algorithm which computes policies for decision problems represented as multi-stage influence diagrams. Our algorithm constructs policies incrementally, starting from a policy which makes no use of the available information. The incremental process constructs policies which includes more of the information available to the decision maker at each step. While the process converges to the optimal policy, our approach is designed for situations in which computing the optimal policy is infeasible. We provide examples of the process on several large decision problems, showing that, for these examples, the process constructs valuable (but sub-optimal) policies before the optimal policy would be available by traditional methods.

## 1 INTRODUCTION

The representational tools which decision analysts and AI practitioners have devised can represent large decision problems. When costs of computation are not taken into account, optimal policies can be determined using dynamic programming [Howard & Matheson, 1984; Shachter, 1986]. When the costs of computation are not negligible, the cost of computing the optimal policy using dynamic programming may be prohibitive.

We have developed an algorithm which can be used to compute policies for large multi-stage decision problems under uncertainty represented as influence diagrams. Our approach is incremental, and uses abstraction. The algorithm is sufficiently general to make use of existing tools for probabilistic reasoning, and has already provided reasonably valuable (but non-optimal) policies for influence diagrams with about $2^{61}$ states.

The algorithm is an extension of the iterative refinement technique presented in [Horsch & Poole, 1996], applied to multi-stage influence diagrams. The refinement is applied to the decision nodes in *random access* ordering (as opposed to the sequential ordering of dynamic programming).

This paper is organized as follows. First we briefly discuss influence diagrams and the decision tree representation of decision functions. Section 2 presents the random access algorithm. Empirical results are presented in Section 3.

### 1.1 INFLUENCE DIAGRAMS

An influence diagram (ID) is a DAG representing a sequential decision problem under uncertainty [Howard & Matheson, 1984]. An ID models the subjective beliefs, preferences, and available actions from the perspective of a single decision maker.

Nodes in an ID are of three types. Random variables, which the decision maker cannot control, are represented by circle shaped *chance* nodes. Decisions, *i.e.*, sets of mutually exclusive actions which the decision maker can take, are represented by square shaped *decision* nodes. The set of outcomes (or actions) which can be taken by a chance node $X$ (or decision node $D$) is specified by $\Omega_X$ (or $\Omega_D$).

The diamond shaped *value* node represents the decision maker's preferences in the form of a value function.

Arcs represent dependencies. A chance node is conditionally independent of its non-descendants given its direct predecessors. The direct predecessors of a decision node will be called *information predecessors*; a value for each of these predecessors will be observed before an action must be taken. The decision maker's preferences are expressed as a function of the value node's direct predecessors. The set of a node's direct predecessors is specified by $\Pi$ subscripted by the node's label.

Dependencies are accompanied by numerical information. There is a conditional probability table associated with every chance node in the form $P(X|\Pi_X)$ (unconditional, if it has no predecessors). The value node $V$ has an associated value function, $V : \Omega_{\Pi_V} \to \Re$, which may be represented as a table.

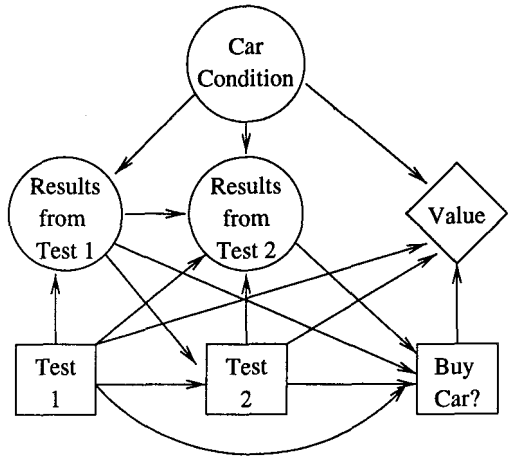

Figure 1: *The Car Buyer Problem, expressed as an influence diagram [Smith, Holtzman, & Matheson, 1993].*

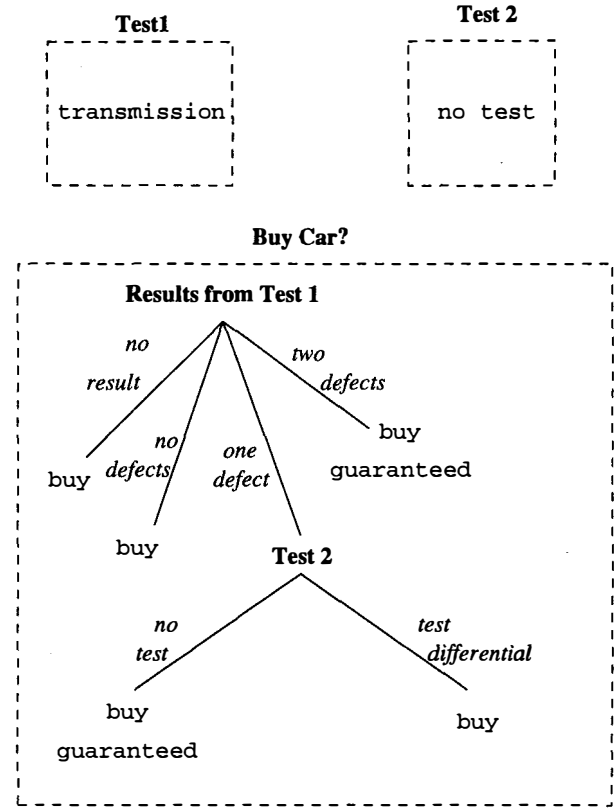

Figure 2: *A policy for the influence diagram in Figure 1. There are three decision trees, one for each decision node:* Test 1, Test 2 *and* Buy Car?.

A policy prescribes an action (or sequence of actions, if there are several decision nodes) for each possible combination of outcomes of its information predecessors. The set $\Omega_{\Pi_D}$ is the set of all possible combinations of values for decision node $D$'s information predecessors. An element in this set will be called an *information state*. A decision function for decision node $D$ is a mapping $\delta : \Omega_{\Pi_D} \to \Omega_D$. A policy for an ID is a set $\Delta = \{\delta_i, i = 1\ldots n\}$ of decision functions, one for each decision node.

An optimal policy maximizes the decision maker's expected value, without regard to the cost of finding such a policy. If computational costs are not negligible, the decision maker's expected value might be maximized by a policy which is not optimal in this sense.

For example, the ID in Figure 1 represents the problem of deciding whether or not to buy a particular car. The decision maker has the option of performing a number of tests to various components of the car. The results of these tests will provide information to the decision to buy the car. The actual condition of the car is not observable directly at the time the decision maker must act, but influences the final value of the transaction. A policy for this problem would indicate which tests to do under which circumstances, as well as a prescription to buy the car (or not) given the results of the tests. Due to space constraints, none of the numerical data required to complete the specification of this problem is shown; this information can be found in [Qi & Poole, 1995; Smith, Holtzman, & Matheson, 1993].

In this paper, IDs are assumed to have chance and decision nodes with a finite number of discrete values. Furthermore, we limit the discussion to IDs with a single value node.

### 1.2 DECISION TREES

Let $D$ be a decision node in an ID. A decision tree $T$ for $D$ is either a leaf labelled by an action $d_j \in \Omega_D$ or a non-leaf node labelled with some $X \in \Pi_D$. Each non-leaf has a child decision tree for every value $x_k \in \Omega_X$. An information predecessor $X \in \Pi_D$ appears at most once in any path from the root to a leaf. Each vertex $X$ in a decision tree has a *context*, $\gamma_X$, defined to be the conjunction of variable assignments on the path from the root of the tree to $X$. The action at the leaf represents the action to be taken in the context of the leaf. Given an information state $w \in \Omega_{\Pi_D}$, there is a corresponding path through a decision tree for $d$, starting at the root leading to a leaf, which is labelled with the prescribed action to be taken in when $w$ is observed.

Note that the context of an action need not contain an assignment for every variable in $\Pi_D$. In this case, the information has not been used in the decision function, even though it is available to the decision maker. In such a situation, a context is said to *cover* a set of information states.

A decision tree represents a decision function. We will refer to the action prescribed by a decision function by $\delta(w)$ for information state $w$, or by $d_l$ if $l$ is a leaf on a given decision tree.

Figure 2 shows three decision trees, one tree for each decision node in the Car Buyer problem (Figure 1). The decision tree for Test 1 is a single leaf, which tells the decision maker to perform the test on the transmission. Since there are no information predecessors for this decision node, this decision tree is complete.

The decision tree for Test 2 tells the decision maker not to perform the test. Note that this decision node has 2 information predecessors. The decision tree does not make use of the available information; every information state is mapped to the action no test.

The decision tree for Buy Car? is a non-trivial tree, using two of four information predecessors. This decision function tells the decision maker to check the result from the first test: if there is *no result* or if there are *no defects*, the decision maker is directed to buy the car. If the result of Test 1 indicates *one defect*, the decision function uses the information from the previous decision Test 2. If the decision to take the second test had been made, the decision maker should buy the car; if the decision maker did not have the second test performed, the car should be bought with a guarantee.

Note that not all of the information is used. A policy which used all of the available information naively would have 96 leaf vertices for Buy Car?; many of these would be logically impossible due to the asymmetries of the problem. The problem is well known for its asymmetry, and the optimal policy can be represented by decision trees very succinctly.

### 1.3 THE SINGLE STAGE ALGORITHM

The single stage information refinement algorithm constructs a decision tree for a influence diagram with a single decision node. The following description is a brief synopsis. The algorithm has been described in more detail in [Horsch & Poole, 1996], and is similar to algorithms described in [Heckerman, Breese, & Horvitz, 1989; Lehner & Sadigh, 1993].

For a given leaf $l$ in a decision tree, its context $\gamma_l$ is extensible if it does not contain all the information variables. We refer to the information variables which are not in the context as *possible extensions*, writing $\xi_l$. A decision tree $t$ can be extended if there is a leaf with an extensible context; otherwise, the tree is called *complete*.

The single stage algorithm can be summarized as follows: A decision tree is extended by removing an extensible leaf $l$ having context $\gamma_l$. This leaf is replaced with new a vertex $X \in \xi_l$. The new vertex $X$ is given a new leaf for every value $x_j \in \Omega_X$. Each leaf has a context $\gamma_j$ which is the assignment of values $(X = x_j) \wedge \gamma_l$. Each leaf out of $X$ will be labelled with an action $d_j \in \Omega_D$. The action $d_j$ is the action which maximizes the expected utility in the new context $\gamma_j = (X = x_j) \wedge \gamma_l$ (this action will be called the *MEV action* for the leaf). The initial tree has one leaf, which is the MEV action to be taken in the empty context.

Other refinement operators are possible. For example, an extension might generate a branch for a particular value of $X$, and summarize the remaining values in a single branch. Determining how and when to use this kind of operator is an avenue for future research.

The sequence of trees created by the procedure is monotonically non-decreasing in expected value. However, the procedure is myopic; there is no guarantee that the expected value will increase with every extension of the tree.

Ideally, an algorithm would choose the extension which maximizes the increase in expected value. The increase in expected value due to a myopic extension can only be determined after the extension has been made. Furthermore, the best extension for a given decision tree can only be determined by extending all the leaf vertices in the tree, and looking at their respective effect on the value of the decision tree.

We use heuristics to avoid computing all myopic extensions for the decision tree. The problem of making the next extension is separated into two parts: the heuristic choice of a leaf, and the strategic choice of an extension for a particular leaf. These tasks are orthogonal [Horsch & Poole, 1996].

We have implemented several heuristics to indicate which leaf to extend. These heuristics are based on domain information available in the influence diagram in terms of probability and expected value. For example, one heuristic chooses to extend the leaf whose context has highest probability. With this heuristic, the most likely situations are explored first. Another of our heuristics looks at the expected value of the possible actions at the leaf; this heuristic orders leaf vertices according to the value of the runner up to the MEV action at every leaf. This is called the *second best action* heuristic, and is based on the intuition that if the value of the second best action is high, it must be close to the value of the best action. In this case, it seems reasonable to explore the context further, since the context may be covering more refined contexts in which the respective actions are very different in value.

Given that a particular leaf has been chosen to be refined, an extension must be chosen for the leaf. There are several strategies which could be used to select one of the possible extensions. For example, a possible extension can be selected at random. The strategy which selects the extension which maximizes the increase in expected utility is called the *maximal* extension strategy. We have also implemented a greedy strategy which chooses the first extension it can find which increases the value of the policy. These strategies and heuristics are discussed in more detail in [Horsch, 1998].

## 2 RANDOM ACCESS REFINEMENT: AN ANYTIME ALGORITHM

In this section, we present an anytime algorithm for computing policies for multi-stage influence diagrams. A policy is represented by a collection of decision trees, one for each decision node in the influence diagram. As in Section 1.3, these decision trees prescribe actions for contexts which may not make use of all the information available to the decision maker. The policy is refined by choosing a leaf from one of these trees and applying a single refinement to the leaf, keeping the rest of the policy fixed.

There is no *a priori* order in which the trees are refined, which is a departure from standard dynamic programming techniques for building an optimal policy. Furthermore, our algorithm always has a policy available, refining it as until the decision maker interrupts the process to act.

While the high level outline of the process is simple, two complications arise in the details. The first is that a deterministic decision tree (as described in Section 1.2) is an inappropriate representation for a decision function in a multi-stage policy which is being refined. The second complication is that for multi–stage decision problems, the refinement may have ramifications for the global policy. Neither of these complications occur for single-stage problems. We describe these complications and our solutions before we present the complete algorithm.

### 2.1 STOCHASTIC DECISION FUNCTIONS

When the decision maker has to act, an unambiguous policy must be available. In single stage problems, an unambiguous policy is represented by a deterministic decision tree. However, during deliberation of multi-stage decision problems, a deterministic decision tree is not a suitable representation of the decision function. Here we describe the problem, and our solution.

The refinement process splits contexts on information predecessors. Consider the situation in which the decision tree for $D_k$ is being refined by splitting on a previous decision $D_i$. Suppose that there are already a decision functions for $D_i$ and $D_k$, and that both are represented as a deterministic decision tree. The split on $D_i$ will not increase the expected value of the decision function for $D_k$, since all but one of the possibilities for $D_i$ would be ruled out by the decision function for $D_i$. The split is still possible, but will have zero effect on the value of the whole policy.

For example, consider Figure 2. If Test 2 were added to the decision function for Buy Car? after the algorithm determined that no test should be performed at Test 2, splitting on Test 2 could not have increased the expected value of the policy. In effect, a deterministic decision function is too committed for the purposes of refining the policy.

To solve this problem, the existing policy can be treated as a stochastic mapping from information state to action. For each context, each available action has an associated probability, representing the belief that future refinement will endorse the action as best in all more refined contexts. This belief is computed by reasoning by cases:

$$P(d_i|\gamma) = pr_i + (1-p)m_i$$

In this expression, $p$ is the probability that no further refinement will occur after the current refinement step (with probability $1-p$, further refinements will occur); $r_i$ is the probability that action $d_i$ will be taken if refinement stops immediately ($r_i = 1.0$ if action $d_i$ is the MEV action in the given context, and 0.0 otherwise); $m_i$ is the probability that action $d_i$ will be taken in any future context derived from the given context.

The parameters $p$ and $m_i$ are assessed by meta-level considerations. We argue that $m_i$ should be close to unity if the expected value of action $d_i$ is relatively high, and close to zero if the expected value is relatively low: one way to realize this intuition is to use $m_i \propto u(d_i|\gamma)$ where $u(d_i|\gamma)$ is the expected value of action $d_i$ in context $\gamma$.

The choice of $p$ is subject to fine tuning (similar to the case of the learning rate in other machine learning algorithms). We argue that $p$ should increase as the policy is refined. Informal experiments indicate that there is a compromise to be made in increasing the value of $p$. If $p$ is increased too slowly or too quickly, the refinement process fails to investigate worthwhile contexts.

A *stochastic decision tree* represents the incomplete decision functions during the random access refinement process. It differs from the decision trees discussed in Section 1.2 only at the leaf vertices. Instead of a single action (the MEV action), the stochastic decision tree labels the leaf $l$ with a probability distribution over the actions $d \in \Omega_{D_k}$, $P(d|\gamma_l)$.

When the refinement process halts, the uncertainty over action in a given context is resolved by setting $p = 1.0$.

### 2.2 THE GLOBAL EFFECTS OF LOCAL REFINEMENT

The second complication is that the refinement process has global effects. For the purpose of refining a particular context $\gamma$ within a decision tree, we assume the remainder of the policy remains fixed. The decision function prescribes an action $d$ for context $\gamma$ already, and the refinement of $\gamma$ may indicate that actions different from $d$ are better for the new contexts derived from $\gamma$[1] The change in the decision function may cause changes to the probability of events after the stage; as well, the change in the decision function may change the expected value of earlier decisions.

---
[1]For refinements to have a positive effect on expected value, a refinement needs to indicate different actions for different contexts.

The changes must be reflected in the decision functions. The expected value of each leaf must be recomputed (we store the expected value at the leaf of the decision tree). As well, we store in our decision trees the probability of each vertex in every context, given the information which precedes it (from the root). These are recomputed as well.

For each internal vertex in all decision trees which follow $D_i$, we need to recompute the posterior probability of the chance node. These can be computed most efficiently using a depth first traversal of each tree, working from $D_{i+1}$ forwards. We observe that changing these probabilities will also have an effect on the expected value of the policy, magnifying the effects of refinement at $D_i$.

After the posterior probabilities have been updated, the expected value of the leaf vertices needs to be recomputed. These are computed starting with the decision tree $D_n$, and working backwards to $D_1$. For each leaf $l$, we need to condition on its context, and recompute the value of action $d_i$ in context $\gamma_l$.

### 2.3 COMPUTING EXPECTED VALUE

To compute expected value, we convert the influence diagram to a Bayesian network, as described in [Shachter & Peot, 1992; Horsch & Poole, 1996]. Briefly, the value node is converted to a chance node; its conditional probability table represents the normalized value function and its complement. We represent decision nodes by chance nodes as well. Initially, the arcs into decision–chance node are dropped, and it is given a uniform probability distribution. When a decision tree is refined, an arc is added in the network if the decision function becomes dependent on an information predecessor. The decision function is installed into the Bayesian network by constructing a conditional probability table consistent with the stochastic decision function and $P(D|\gamma_l)$ at each leaf $l$.

Using this transformation, expected utility can be computed by making a query to the network. The query $P(D|v\gamma)$ gives MEV action for decision node $D$ a given context, where $v$ is the value of the utility–chance node $V$. Note that $\gamma$ must be consistent with $v$ before this query is made; in our implementation, we check that $P(v|\gamma)$ is non-zero before we query for the MEV action. To find the expected value of an action $d$ in a given context $\gamma$, we make the query $P(V|d\gamma)$. As a result, each time a MEV action is computed, 3 queries are made to the network.

```
procedure Random Access Refinement
    Input:
        Multi-stage influence diagram with decision nodes
            D_1, ..., D_n
    Output:
        Policy Δ = {δ_1, ..., δ_n}, a set of decision trees

    For each D_i, initialize δ_i as a single leaf
    Do {
        Choose an extensible decision tree δ_i
        Choose a leaf from δ_i
        Replace the leaf with an extension
        Install the modified decision function
        Update the global policy
    } Until (stopping criteria are met or policy is complete)
    Return the policy
```

Figure 3: *The random access refinement algorithm.*

### 2.4 THE RANDOM ACCESS REFINEMENT ALGORITHM

The high level description of the algorithm is given in Figure 3. The algorithm is discussed briefly step by step.

**Initialization:** The initialization process considers each decision node in order $D_n, \ldots, D_1$. For each decision node, the probability distribution $P(D_i)$ is determined for the empty context. This step requires three queries to the Bayesian network for each decision node.

**Choosing a decision function to refine:** We maintain a priority queue of extensible leaf vertices, ordered by heuristic value. The queue contains pairs $(D_i, l)$ where $D_i$ is a decision node, and $l$ is a leaf on the decision tree for $D_i$. Thus, the heuristic value assigned to a leaf determines not only the order in which the leaf vertices for a single tree are extended, but also the the order in which the decision functions are refined. As a result, decision functions are refined in order of the heuristic importance of the refinement, rather than a predetermined sequence. The heuristics discussed in Section 1.3 can be used for this dual purpose.

**Extending a given leaf:** As in the single stage algorithm, an extension is chosen for a given leaf. This can be done by one of the strategies described briefly in Section 1.3.

**Updating the global policy:** Each decision tree $D_{i+1}, \ldots, D_n$ has its observation probabilities updated: for each vertex $X$, recompute $P(X|\gamma_X)$. The chance node representing the decision in the Bayesian network is changed to match the update.

Each decision tree $D_n, \ldots, D_1$ has its expected value updated. For each leaf vertex, a single query for $P(D|v\gamma)$ will provide a vector of $m_i$ values, from which we can compute $P(D_i|\gamma)$ as in Section 2.1. The query $P(V|d^*\gamma)$ will give the expected value of the best action. Finally, the chance node representing the decision in the Bayesian network is changed to match the update.

## 2.5 COMPLEXITY

We can analyze the cost of this procedure as follows. Suppose a decision node has $n$ information predecessors, each with at most $b$ values. To find a maximal extension for a single leaf requires $O(b(n - k))$ expected value computations, where $k$ is the number of internal vertices already in the context for the leaf.

An update of the global policy requires one computation of posterior probability for each internal vertex and 2 expected value computations for each leaf. In the worst case all the stages have probabilities and expected values updated. The total number of leaf nodes on all the trees is $O((b-1)N + D)$, where $N$ is the number of refinements which have been made in total, and $D$ is the number of decision nodes in the influence diagram. The total number of internal vertices in all the decision trees is $O((b-1)N + D)$.

Each computation of expected value is equivalent to a query in a Bayesian network [Shachter & Peot, 1992]. Thus, the total cost, in terms of the number of queries to a Bayesian network, of the a single refinement and update is $O(b(n - k) + 3((b-1)N + D))$.

In the worst case, the procedure requires $O(b^{n+1})$ queries just for the refinements for a complete policy. In the worst case, the updates after each refinement add $O(b^{2n})$ total queries updating the policy after each refinement. This is substantially more effort than is required by an exhaustive enumeration of the state space; however, for large state spaces, a policy is available for use by the decision maker with much smaller cost than the limit of a complete policy.

The next section applies the random access refinement algorithm to some large decision problems, demonstrating that the process constructs valuable policies at a fraction of the cost of computing the optimal policy using exhaustive enumeration.

## 3 EMPIRICAL RESULTS

The random access refinement process is intended to find valuable policies with a relatively small investment of computational resources. A number of large influence diagrams were constructed to demonstrate that the algorithm does achieve this intention. The influence diagrams are identical in topology, but the conditional probabilities vary. The problems have a real interpretation, in contrast to randomly generated problems. The purpose of running the algorithm on slightly varying problems is to demonstrate the effect of variations in the problem on the performance of the algorithm.

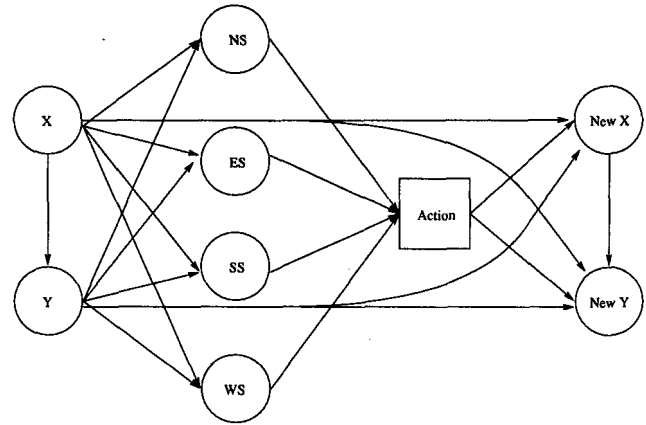

Figure 4: *A influence diagram fragment, showing a single stage for variations of the maze walker problem. The problems solved in this paper iterate this structure ten times.*

### 3.1 THE PROBLEMS

The decision problems are based on the model of an agent traversing a maze. The mazes consist of walls and open space, and are represented by square tiles whose size correspond to the agent's single step. The agent has five available actions: it can move a single step in any of the four compass directions *N, S, E, W*, or stay in place. The agent has four sensors *NS, ES, SS, WS*, one in each compass direction.

The agent can only detect walls (with or without noise); the agent's position is not directly observable. The goal of the agent is to arrive at a specified location in the maze.

The problem of choosing an action can be represented by an influence diagram; the representation imposes a finite structure on the problem, namely that the agent is limited to a fixed number of actions. A single stage is shown in Figure 4. The four sensors are directly connected to the decision node. The two state variables affect the sensors directly, but are themselves not directly observable by the agent. In principle, the single stage can be repeated any number of times; no-forgetting arcs connect the maze walker's previous sensors and actions to the the current action. In the figure, the no-forgetting arcs have not been drawn.

The probabilistic information required by this influence diagram forms the agent model. Sensors can be modelled with the conditional probability distributions $P(NS|X,Y)$, *etc*. Actuators can be modelled by the conditional probability distributions $P(NewX|X,Y,Action)$ and $P(NewY|X,Y,Action,NewY)$.

Four agent models were used in this test. These correspond to two sensor models: perfect and noisy; and two actuator models: perfect and noisy. The perfect sensors always detect a wall when there is one, and never detect a wall when there isn't one. The noisy sensor model has probability 0.9 that a wall is correctly detected, and 0.05 that a wall is detected when no wall is there. The perfect actuators always put the agent in the correct square for a given action. The noisy actuator model depends on adjacent walls and obstacles. The agent ends up in the right place for a given action with a probability of about 0.89, and with probability about 0.089, the agent fails to move. The noisy actuator has a very small probability (about 0.01) of moving to an incorrect adjacent square.

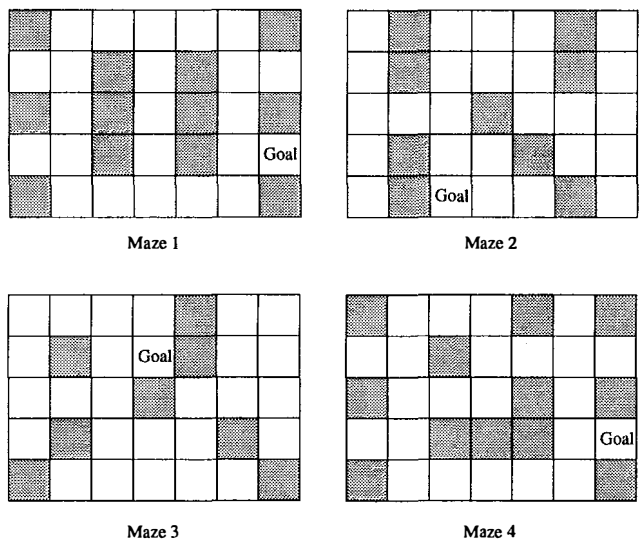

Figure 5: *The mazes for the maze walker problem. The shaded tiles are obstacles, and there are walls around the perimeter of the maze.*

The value function is not shown in the ID fragment. It depends only on the position of the agent in the final stage, and puts full value (1.0) on being at the goal, and zero elsewhere.

The mazes used in our experiments are shown in Figure 5 (Maze 1 is an example from [Littman, Cassandra, & Kaelbling, 1995]). In our experiments, the agent is allowed ten stages to reach the goal, which makes it possible to reach the goal from each starting position. Using 10 stages, the tenth decision node has 49 direct predecessors.

Maze 1 has a simple policy which guides the perfect agent to the goal from each possible starting position. The policy guides the agent south whenever possible, or otherwise east whenever possible. If neither south nor east is possible, the agent moves west, if possible, and otherwise stays in place. This decision function is repeated for the first 8 stages. The final two steps of the policy direct the agent north one step and east one step. This policy has an expected value of 1.0, and can be represented by 8 decision trees which use 3 internal vertices each, followed by two decision trees which need no internal vertices.

Maze 2 has an ambiguity which cannot be resolved by following a path to the goal. An optimal policy can guide the perfect agent to the goal position from 24 of the 25 starting positions of this maze, for a maximum expected value of 0.96. We estimate that an optimal policy for the perfect agent in this maze can be represented by 10 decision trees using a total of about 30 internal vertices.

We do not have optimal policies for Mazes 3 and 4, but all the ambiguities in these mazes can be resolved along a path to the goal, *i.e.*, there exist policies which guide the perfect agent to the goal from all starting positions; these policies have expected value of 1.0. We estimate that the optimal policies can be represented by 10 decision trees using between 20 and 30 internal vertices in total.

The optimal policies for the agents with imperfect sensors or actuators are unknown; the value of the optimal policy depends in part on the difficulty of the maze.

### 3.2   THE RESULTS

The random access refinement algorithm was applied to these problems. The *second best action* heuristic was used to select leaf vertices to extend, and the *maximal* extension strategy was used to extend each leaf. The algorithm had 20 extensions in total allocated for each problem. Note that this resource limit excludes the optimal policy for all the mazes. The average run time on a SPARC Ultra-2 for these problems was 73 minutes.

Figure 6 shows 4 datasets, corresponding to the variations of the agent model navigating Maze 1. The x-axis measures computational costs, in terms of the number of posterior probabilities and expected values computed (queries to the Bayesian network). The y-axis measures expected value of each policy. Each point on a curve represents the value of a policy in the sequence of policies constructed by the algorithm. The first policy is the same for each of the problems, and represents the value of acting randomly before any deliberation has occurred.

For the perfect agent, the algorithm does not find the optimal policy using the allotted resources, but levels off at an expected value of 0.869565 after 2280 steps. The policy guides the agent to the goal from 20 of the 23 starting positions. This is roughly what one might expect, given that the optimal policy uses 24 internal vertices, and the algorithm was given resources to include only 20 internal vertices. The error here is 13% from optimal. We do not currently know whether the refinement process will find an optimal policy in reasonable time.

The curves in Figure 6 give an indication of how the conditional probabilities underlying the agent model affect the performance profile. When the probabilities are very sharp, and a few states contain most of the probability mass (as in the case of the perfect agent), the increases tend to be steep and plateaus are common. As the probability mass of is distributed over many more states (as in the agent with noisy sensors and noisy actuators), the increases tend to be less steep, and the plateaus shorter. These curves are typical.

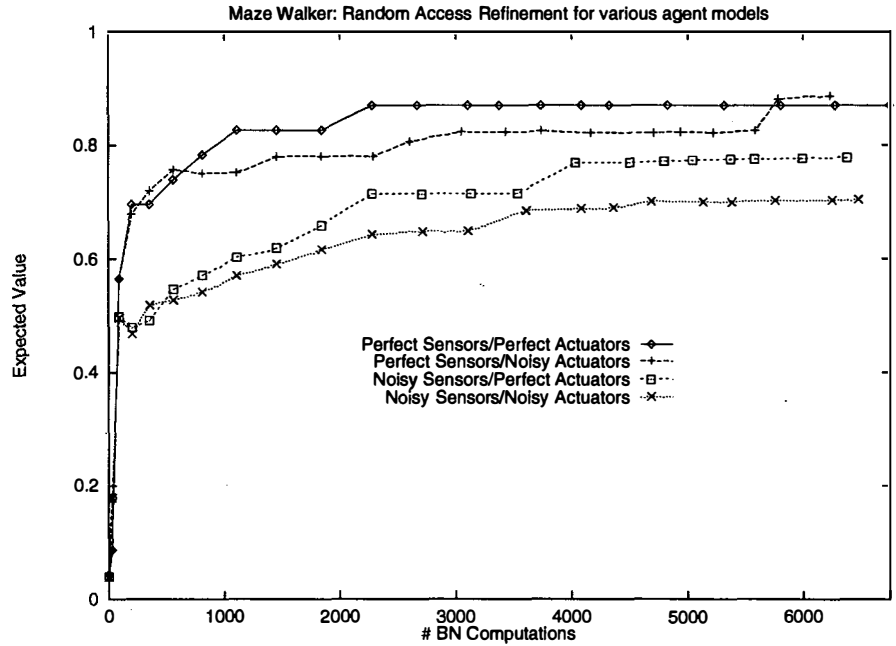

Figure 6: *The performance of the random access refinement algorithm using various agent models for the Maze Walker problem (Maze 1).*

Table 1 summarizes the performance of the various agents in the various mazes. The error for the perfect agent for all four mazes is 13%, 20%, 30%· and 9%, respectively. For comparison, exhaustive enumeration would require more than $2^{60}$ queries to compute complete policies. [2]

For some of the agent models, the algorithm produces policies which decrease in value (for example, in the range of 0 to 500 queries in Figure 6). This behaviour is the result of making a local refinement when the commitment to the current policy is weak. The refinement takes advantage of the relatively high probability of a non-MEV action. When the effects of the refinement are made global, the non-MEV action drops in probability, and any action which was based on the non-MEV action will drop in value. This drop in value is temporary, and further refinement, stronger commitment, and global updates correct for the decrease.

The curve for the agent with noisy actuators and perfect sensors also shows slight decreases in expected value in the range of 3500 to 5500 steps, followed by slight increases. This decrease has the same explanation as the more dramatic decreases observable earlier in the sequence. The ef-fect is smaller since the commitment to the MEV action is stronger.

Also of note is the fact that the algorithm is able to find a policy for the agent with noisy actuators and perfect sensors which exceeds the value of the best policy for the perfect agent. This behavior is due to the heuristics used by the algorithm. In the case of the perfect agent, the heuristic chose to examine a certain set of contexts first. The noisy actuators in the other agent gave a different heuristic value to the contexts.

Some of the variations on the Maze Walker have a relatively large number of impossible information states; policies which summarize a large subspace of the information set can exploit these asymmetries, by not refining impossible contexts. Furthermore, if there is a subset of information states which cover most of the probability mass, it is possible to summarize a large portion of the state space by examining the most likely observations. Druzdzel [1994] argues that it is common for a few states to cover a large portion of the total probability mass in a joint probability distribution. Thus it seems reasonable to expect that policies which contain a small number of contexts will achieve fairly high value. The data presented in this paper support this expectation.

Finally, it is important to acknowledge that the space of IDs is very large, and the set of problems treated in this section is a small sample from a highly restricted subclass of IDs. The evidence in this section suggests that there exist large problems for which random access refinement can find poli-

---

[2] To get an idea of the scale of this number: the figure is about 10 cm wide; at this scale, $2^{60}$ queries is approximately 15 light-hours to the right. It would take about 23 billion years to compute according to the average reported above.

| Agent Model Sensor/Actuator | Best Policy | Steps |
|---|---|---|
| Perfect/Perfect | 0.8696 | 2280 |
| Perfect/Noisy | 0.8874 | 6236 |
| Noisy/Perfect | 0.7767 | 6374 |
| Noisy/Noisy | 0.7045 | 6474 |

Maze 1

| Agent Model Sensor/Actuator | Best Policy | Steps |
|---|---|---|
| Perfect/Perfect | 0.7692 | 4962 |
| Perfect/Noisy | 0.5159 | 5355 |
| Noisy/Perfect | 0.5887 | 5838 |
| Noisy/Noisy | 0.4703 | 5775 |

Maze 2

| Agent Model Sensor/Actuator | Best Policy | Steps |
|---|---|---|
| Perfect/Perfect | 0.7037 | 4522 |
| Perfect/Noisy | 0.5452 | 5581 |
| Noisy/Perfect | 0.6169 | 6079 |
| Noisy/Noisy | 0.4933 | 5799 |

Maze 3

| Agent Model Sensor/Actuator | Best Policy | Steps |
|---|---|---|
| Perfect/Perfect | 0.9130 | 4564 |
| Perfect/Noisy | 0.6511 | 6219 |
| Noisy/Perfect | 0.6760 | 5319 |
| Noisy/Noisy | 0.6270 | 6162 |

Maze 4

Table 1: *A summary of the best policies found by the random access refinement algorithm applied to several large decision problems. The optimal policy for the perfect agents is known to have expected value 1.0 for mazes 1, 3 and 4, and 0.96 for maze 2. The optimal policy for these problems could be computed using dynamic programming, requiring about $2^{60}$ steps.*

cies which are reasonably valuable policies using reasonable amounts of computational resources. These problems are too large to solve using traditional methods.

## 4 RELATED WORK

The information refinement approach is closely related to learning classification trees in machine learning (*e.g.*, [Quinlan, 1986]). Heckerman *et al.*[1989] discusses an algorithm which constructs policies in a similar manner. Their interest is in representing a policy which can be used effectively by the decision maker on-line. The costs of building the decision tree are not taken into account; the costs of using the decision tree are compared to the cost of other on–line approaches.

Lehner and Sadigh [1993] also discusses the issue of compiling a decision problem into a *situation-action* tree. They do not emphasize computational cost; their goal is to take a complex problem and create rules for use by human decision makers. They determine the best decision tree of a certain size, regardless of the cost of computing them.

Zhang & Boerlage [1995] simplify decision problems by removing inconsistent information states and "insignificant details" before constructing a policy for the problem. The significance of the details in the information state is measured in terms of the effects of the information state on the posterior probabilities of (unobservable) state variables.

Horvitz and Klein [1993] describe a decision theoretic approach to categorization based on utility. By aggregating states with similar utility values, and actions with similar values, decision models can be simplified for increased efficiency. Poh and Horvitz [1993] presents a greedy approach to exploring how random variables in a decision model might be refined, *i.e.*, how they can be given a more fine–grained set of values, to increase the utility of a decision. This work is intended to automate some of the effort that a decision analyst would put into reframing a decision problem, and deals with the refinement problem on a lower level than information refinement.

Information refinement is closely related to "input generalization" which is used to help deal with large state spaces in reinforcement learning. Chapman and Kaelbling [1991] adapt the Q-learning algorithm for large input spaces by using a decision tree in place of the table to represent the $Q$-function. The decision tree is extended by "splitting" the function on significant input bits, as determined by tests for perceptual and value significance.

## 5 CONCLUSIONS

We have described an anytime algorithm for information refinement in multi-stage decision problems represented as influence diagrams. The process builds a stochastic decision tree for each decision node in the influence diagram. Each tree is initialized to be a single leaf labelled with the best action to perform without using any of the available information. A leaf is chosen heuristically, and is replaced with an extension. A probability distribution is imposed over the actions in the policy, which is a subjective assessment of the probability that any particular action will be carried out once the anytime refinement process is halted. The global effects of the refinement are propagated through the decision trees of the policy; probabilities are recomputed for decision trees following the refinement, and all leaf vertices are recomputed in all the decision trees.

The procedure is very expensive asymptotically, and it is possible to construct an influence diagram for which the anytime algorithm will construct policies which are have no more than 50% of the expected value of the the optimal policy as long as no contexts are complete. An example of this kind of influence diagram has only uniform probability distributions and a value function in the form of the parity function on its inputs.

The results shown in this paper demonstrate that information refinement constructs reasonably valuable policies for large decision problems using reasonable amounts of computational resources. For some of the influence diagrams treated in this paper, no optimal policy is known. These problems are too large to enumerate the information space exhaustively.

### Acknowledgements

The authors would like to thank the reviewers for helpful suggestions, and also Brent Boerlage of Norsys Software Corp. for advice and support in the use of the Netica API as our Bayesian network engine, and for many discussions. The second author is supported by the Institute for Robotics and Intelligent Systems, Project IC-7, and the National Sciences and Engineering Council of Canada Operating Grant OGPOO44121.